\pdfoutput=1
\documentclass[final]{cvpr}

\usepackage{times}
\usepackage{epsfig}
\usepackage{graphicx}
\usepackage{amsmath}
\usepackage{amssymb}

\usepackage[group-separator={,}]{siunitx}
\usepackage{xcolor}
\usepackage{booktabs}

\newcommand{\abs}[1]{\left|#1\right|}

\usepackage[pagebackref=true,breaklinks=true,colorlinks,bookmarks=false]{hyperref}

\begin{document}

%%%%%%%%% TITLE
\title{Yelp Review Rating Prediction: \\
	Machine Learning and Deep Learning Models}

\author{Zefang Liu\\
Georgia Institute of Technology\\
North Avenue, Atlanta, GA, USA\\
{\tt\small \href{mailto:liuzefang@gatech.edu}{liuzefang@gatech.edu}}
}

\maketitle

%-------------------------------------------------------------------------------
%%%%%%%%% ABSTRACT
\begin{abstract}
   We predict restaurant ratings from Yelp reviews based on Yelp Open Dataset. Data distribution is presented, and one balanced training dataset is built. Two vectorizers are experimented for feature engineering. Four machine learning models including Naive Bayes, Logistic Regression, Random Forest, and Linear Support Vector Machine are implemented. Four transformer-based models containing BERT, DistilBERT, RoBERTa, and XLNet are also applied. Accuracy, weighted $ F_1 $ score, and confusion matrix are used for model evaluation. XLNet achieves 70\% accuracy for 5-star classification compared with Logistic Regression with 64\% accuracy.
\end{abstract}

%%%%%%%%% BODY TEXT
%-------------------------------------------------------------------------------
\section{Introduction}

Online reviewing for businesses becomes more and more important nowadays, where customers can publish their reviews for businesses, and other potential customers or shop owners can view them. Positive feedback from customers may prosper the store businesses, while negative one could have opposite consequences. Yelp, one of the largest company founded in 2004 for publishing crowd-sourced reviews about businesses, provides one open dataset, Yelp Open Dataset \cite{Yelp}, which has tremendously many data about businesses, reviews, and users. Such dataset has been proven to be a good material for personal, educational, and academic purposes.

Among multiple tasks on the Yelp Open Dataset, predicting ratings for restaurants based their reviews is one of fundamental and important tasks. This task can help Yelp classify reviews into proper groups for its recommendation system, detect anomaly reviews to protect businesses from malicious competitions, and assign rating to texts automatically.

Yelp review rating prediction can be done in multiple ways, such as sentiment analysis and 5-star rating classification. In this paper, we will focus on rating prediction for restaurants based only on their review texts. This task can be viewed as a multiclass classification problem, where the input is the textual data (reviews), and output is the predicted class (1-5 stars). We will apply both machine learning and deep learning models. After analyzing data distribution, splitting datasets, and extracting features, we will use four machine learning methods, including Naive Bayes, Logistic Regression, Random Forest, and Linear Support Vector Machine (SVM) \cite{ESL}. Then we will focus on four transformer-based models, including BERT \cite{BERT}, DistilBERT \cite{DistilBERT}, RoBERTa \cite{RoBERTa}, and XLNet \cite{XLNet}, where several different architectures will be tried with hyperparameter tuning. This project is done on \href{https://colab.research.google.com/}{Google Colab}, where multi-processors and GPUs are available. The code is publicly available at GitHub \footnote[1]{\url{https://github.com/zefang-liu/yelp-review}}.

\section{Related Work}

In past years, many projects have been done on the Yelp Open Dataset. The data distribution has been analyzed thoroughly in \cite{hajas2014analysis,cui2015evaluation}. The review rating prediction has also been performed. Multiple feature generation methods and several machine learning models including Naive Bayes, Logistic Regression, Support Vector Machine (SVM), and Gaussian Discriminant Analysis have been used for this classification task in \cite{fan2014predicting,elkouri2015predicting,asghar2016yelp,yu2017identifying}. One reported best accuracy for the 5-star classification is 64\% on the testing set in \cite{elkouri2015predicting}. Some deep learning model such as Neural Network, Recurrent Neural Network (RNN), Long Short-Term Memory (LSTM), and Bidirectional Encoder Representations (BERT) were also applied in \cite{perez2017predicting,liu2020sentiment}.

%-------------------------------------------------------------------------------
\section{Dataset and Preprocessing}

We will use \href{https://www.yelp.com/dataset}{Yelp Open Dataset} \cite{Yelp} in this paper. This Yelp dataset is a subset  of Yelp business data, which supplies a good material for academic and research projects. Ideas and methods from machine learning, deep learning, and natural language processing (NLP) can be built and tested on this dataset. This dataset includes \SI{8021122}{} reviews from \SI{209393}{} businesses in 10 metropolitan areas. These data are structured in JSON files, including business, review, user, check-in, tip, and photo data. We will only use business and review data.

%-----------------------------------------------------------
\subsection{Data Preparation}

In order to obtain more accurate rating prediction, we will only use businesses data in restaurant category. We obtain \SI{63944}{} restaurants business IDs from \texttt{business.json}. Then, their \SI{5055992}{} reviews are extracted from \texttt{review.json}. Some samples of data are shown in the Table \ref{tab 2-0}. Rating stars are between 1 star (worst) and 5 stars (best).

\begin{table}[!ht]
	\centering
	\caption{Data Samples}
	\vspace{4pt}
	\begin{tabular}{cc}
		\toprule
		Text & Stars \\
		\midrule
		I love Deagan's. I do. I really do. The atmosp... & 5 \\
		Dismal, lukewarm, defrosted-tasting "TexMex" g... & 1 \\
		Oh happy day, finally have a Canes near my cas... & 4 \\
		This is definitely my favorite fast food sub s... & 5 \\
		Really good place with simple decor, amazing f... & 5 \\
		\bottomrule
	\end{tabular}
	\label{tab 2-0}
\end{table}

%-----------------------------------------------------------
\subsection{Data Distribution}

\begin{figure}[!ht]
	\centering
	\includegraphics[width=0.8\linewidth]{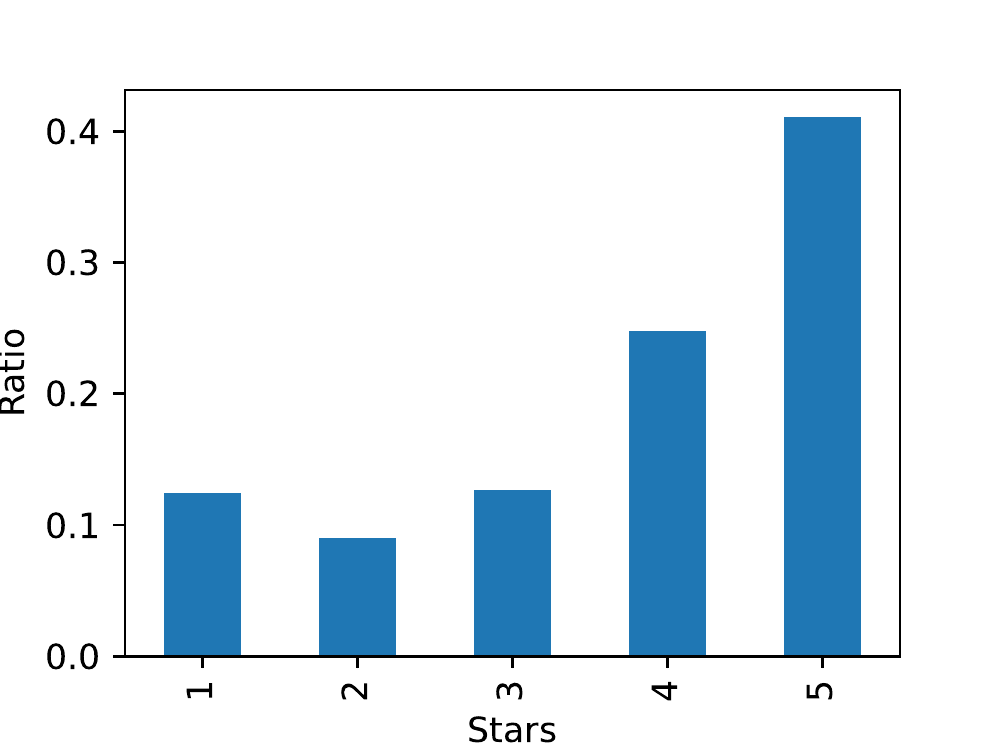}
	\caption{Stars Distribution}
	\label{fig 2-1}
\end{figure}

\begin{figure}[!ht]
	\centering
	\includegraphics[width=\linewidth]{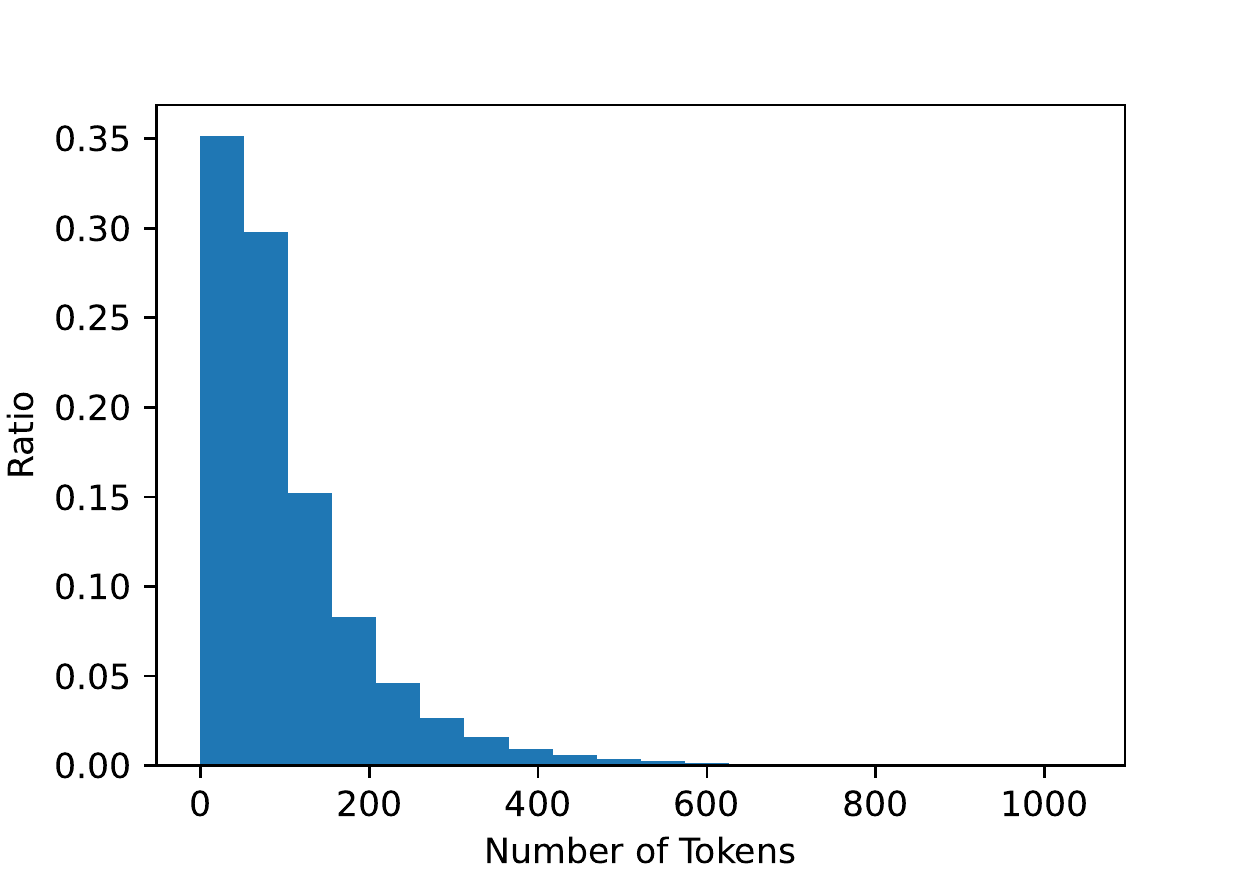}
	\caption{Number of Tokens Distribution}
	\label{fig 2-2}
\end{figure}

The numbers of reviews from each rating category are not equal. Such data distribution is shown in the Table \ref{tab 2-1} and Figure \ref{fig 2-1}. We can find the data are skewed towards two polarities, and majority (65.9\%) reviews are in 4 and 5 stars. This imbalanced distribution was verified in \cite{cui2015evaluation}. Besides, the numbers of tokens in each reviews are also imbalanced. The histogram of the number of tokens is shown in the Figure \ref{fig 2-2}. The distribution shows that 73.1\% reviews are no longer than 128 tokens, and 92.8\% reviews are no longer than 256 tokens.

\begin{table}[!ht]
	\centering
	\caption{Data Distribution}
	\vspace{4pt}
	\begin{tabular}{ccc}
		\toprule
		Stars & Number of Reviews & Percentage \\
		\midrule
		1 & \SI{628044}{} & 12.42\% \\
		2 & \SI{456590}{} & 9.03\% \\
		3 & \SI{639748}{} & 12.65\% \\
		4 & \SI{1254099}{} & 24.80\% \\
		5 & \SI{2077511}{} & 41.09\% \\
		\bottomrule
	\end{tabular}
	\label{tab 2-1}
\end{table}

%-----------------------------------------------------------
\subsection{Dataset Splitting}

In order to deal with the imbalanced raw data, we need to reconstruct a balanced training dataset. Thus, \SI{1250000}{} reviews are resampled from the raw data for a training dataset, where \SI{250000}{} reviews are from each category (1-5 stars). Besides, both validation and testing datasets have \SI{250000}{} reviews, but they roughly follow the imbalanced distribution of the raw data.

%-----------------------------------------------------------
\subsection{Feature Extraction}

With the purpose of using the machine learning models, we firstly build features by converting textual documents to numerical data (matrices). Some detailed text preprocessing pipelines were given in \cite{liu2020sentiment,asghar2016yelp}. We will use vectorizers from \texttt{scikit-learn} \cite{scikit-learn} with the following settings:
\begin{itemize}
	\item Implementing binary and integer version Count Vectorizer and Tf-Idf Vectorizer for word representation,
	\item Using unigram and bigram,
	\item Setting minimum document frequency as 5,
	\item Converting all characters to lowercase,
	\item And removing English stop words.
\end{itemize}

Count Vectorizer can convert text documents to a token counts matrix, while Tf-Idf Vectorizer will used tf–idf instead of token counts. Tf-idf is a product of term-frequency $ \text{tf}(t,d) $ and inverse document-frequency $ \text{idf}(t) $:
\begin{equation}
\text{tf-idf}(t,d) = \text{tf}(t,d) \times \text{idf}(t) ,
\end{equation}
where $ \text{tf}(t,d) $ is the frequency of term $ t $ in the document $ d $, and $ \text{idf}(t) $ is defined by
\begin{equation}
\text{idf}(t) = \log{\frac{n}{1+\text{df}(t)}} ,
\end{equation}
where $ n $ is the total number  of documents in this document set, and $ \text{df}(t) $ is the number of documents in the document set that contain term $ t $. For the binary version vectorizer, all non-zero counts are set to 1. Sparse matrices will be returned from these vectorizers.

In order to compare the performances of these vectorizers, we test their generated feature matrices with logistic regression. The comparison results on the validation set are shown in the Table \ref{tab 3-1}. The results show that the integer version Tf-Idf vectorizer has the best performance on the validation set, which will be used for machine learning models.

\begin{table}[!ht]
	\centering
	\caption{Classification Metrics for Different Vectorizers on the Validation Set}
	\vspace{4pt}
	\begin{tabular}{ccc}
		\toprule
		Vectorizer & Accuracy & $ F_1 $ Score \\
		\midrule
		Count (Integer) & 0.6321 & 0.6369 \\
		Count (Binary) & 0.6285 & 0.6293 \\
		Tf-Idf (Integer) & 0.6387 & 0.6431 \\
		Tf-Idf (Binary) & 0.6358 & 0.6420 \\
		\bottomrule
	\end{tabular}
	\label{tab 3-1}
\end{table}

For the transformer-based models (e.g. BERT), we use textual data directly without handmade features. More further descriptions will be presented in the transformer-based models section.

%-------------------------------------------------------------------------------
\section{Experiments and Results}

In this section, we will apply machine learning and deep learning models for rating prediction on Yelp review. Evaluation metrics will be introduced first, then four machine learning models and four transformer-based models will be implemented for this task. Comparisons and conclusions will be derived at the end.

%-----------------------------------------------------------
\subsection{Evaluation Metrics}

In order to evaluate machine learning and deep learning models in a fair way, we will use the following evaluation metrics:
\begin{itemize}
	\item Accuracy classification score: the fraction of correctly classified samples.
	\item Weighted averaged $ F_1 $ score \cite{F1}:  
	\begin{equation}
	\dfrac{1}{\sum_{l \in L} \abs{\hat{y}_l}} \sum_{l \in L} \abs{\hat{y}_l} F_1 \left( y_l, \hat{y}_l \right), 
	\end{equation}
	where $ L $ is the set of labels, $ y_l $ is the subset of true $ y $ with label $ l $, $ \hat{y}_l $ is the subset of predicted $ \hat{y} $ with label $ l $, and 
	\begin{equation}
	F_1 = \frac{2 \times \text{precision} \times \text{recall}}{\text{precision} + \text{recall}} .
	\end{equation}
	\item Confusion matrix \cite{Matrix}: a matrix $ C $  to evaluate the accuracy of a classification, where $ C_{i,j} $ is the ratio of the number of observations in group $ i $ and predicted to be in group $ j $ over the number of total observations in group $ i $.
\end{itemize}
One overall list of models and their evaluation results are shown in Table \ref{tab 4-1}. The corresponding training time for them on Google Colab is also presented in the Table \ref{tab 4-2}, where running time may vary significantly due to computation resource fluctuation. More details about model implements and experiment results are provided in the following subsections.

\begin{table}[!ht]
	\centering
	\caption{Model Evaluations on the Testing Set}
	\vspace{4pt}
	\begin{tabular}{ccc}
		\toprule
		Model & Accuracy & $ F_1 $ Score \\
		\midrule
		Naive Bayes & 0.6150 & 0.6222 \\
		Logistic Regression & 0.6407 & 0.6454 \\
		Random Forest & 0.5954 & 0.5870 \\
		Linear SVM & 0.6199 & 0.6043 \\
		\midrule
		BERT (base, uncased) & 0.6911 & 0.6963 \\
		BERT (base, cased) & 0.6971 & 0.7013 \\
		BERT (large, cased) & 0.7004 & 0.7050 \\
		DistilBERT (base, uncased) & 0.6847 & 0.6897 \\
		DistilBERT (base, cased) & 0.6944 & 0.6985 \\
		RoBERTa (base) & 0.7029 & 0.7080 \\
		XLNet (base, cased) & 0.7044 & 0.7087 \\
		\bottomrule
	\end{tabular}
	\label{tab 4-1}
\end{table}

\begin{table}[!ht]
	\centering
	\caption{Model Training Time on Google Colab with 4 Concurrent Workers and 1 GPU (run-time format in hour:minute:second)}
	\vspace{4pt}
	\begin{tabular}{cc}
		\toprule
		Model & Training Time \\
		\midrule
		Naive Bayes & 00:00:05 \\
		Logistic Regression & 00:12:42\\
		Random Forest & 00:58:42 \\
		Linear SVM & 00:00:35 \\
		\midrule
		BERT (base, uncased) & 05:36:05 \\
		BERT (base, cased) & 02:38:37 \\
		BERT (base, large) & 05:02:00 \\
		DistilBERT (base, uncased) & 02:54:52 \\
		DistilBERT (base, cased) & 01:21:47 \\
		RoBERTa (base) & 05:32:44 \\
		XLNet (base, cased) & 07:11:19 \\
		\bottomrule
	\end{tabular}
	\label{tab 4-2}
\end{table}

%-----------------------------------------------------------
\subsection{Machine Learning Models}

In the first part of experiment, we apply four machine learning models, i.e. Naive Bayes, Logistic Regression, Random Forest, and Linear Support Vector Machine (SVM), for the Yelp review rating prediction task. Classifiers from \texttt{scikit-learn} \cite{scikit-learn} are used.

%-----------------------------------------------------------
\subsubsection{Naive Bayes}

\begin{figure}[!ht]
	\centering
	\includegraphics[width=\linewidth]{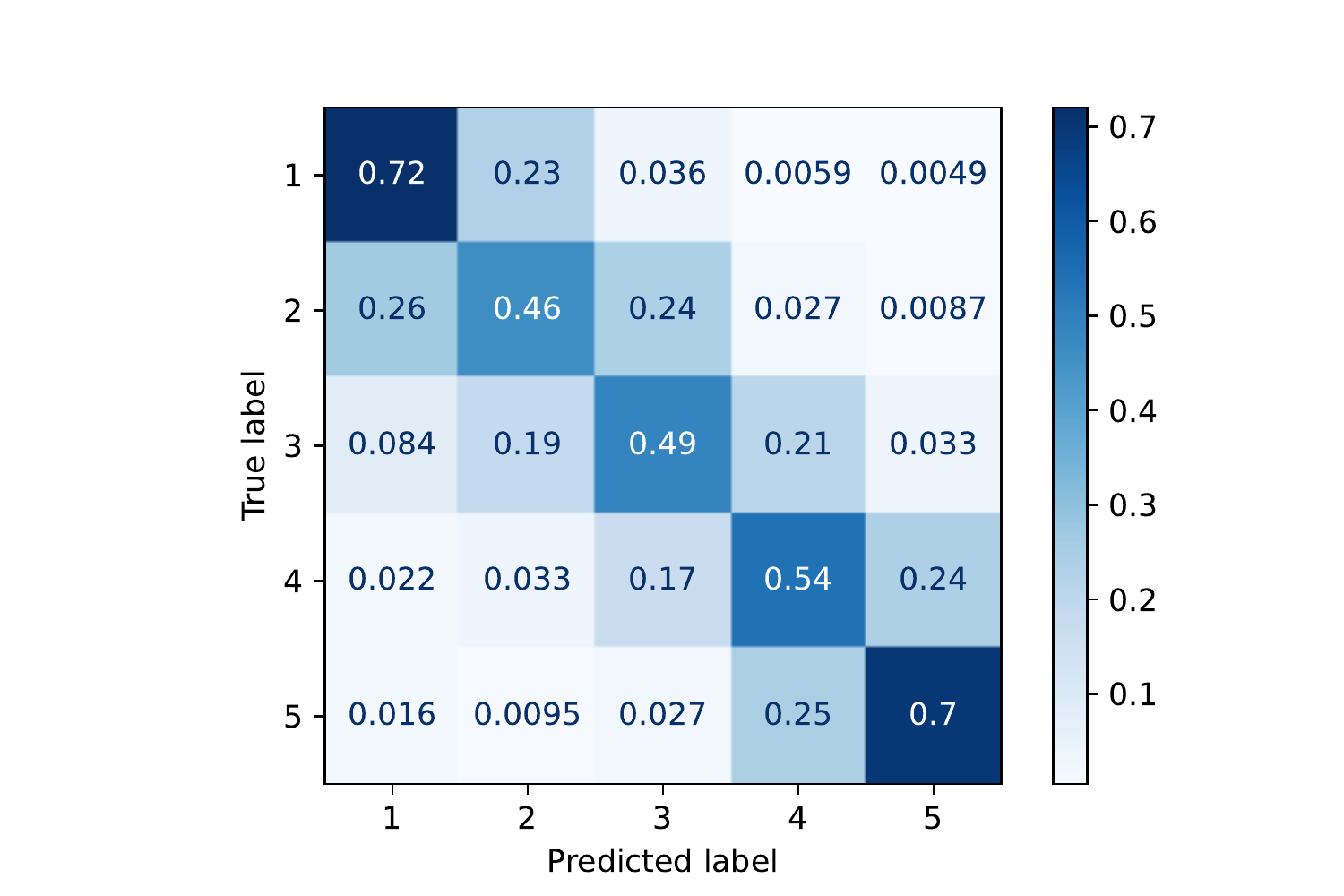}
	\caption{Confusion Matrix for Naive Bayes on the Testing Set}
	\label{fig 4-0}
\end{figure}

The first applied machine learning model is Naive Bayes classifier \cite{ESL} for multinomial models (\texttt{MultinomialNB}). The accuracy and weighted $ F_1 $ score on the testing set are shown in the Table \ref{tab 4-1}, and the confusion matrix is shown in the Figure \ref{fig 4-0}. Naive Bayes model is convenient to be implemented with short running time, which can be viewed as the benchmark for further models.

%-----------------------------------------------------------
\subsubsection{Logistic Regression}

\begin{figure}[!ht]
	\centering
	\includegraphics[width=\linewidth]{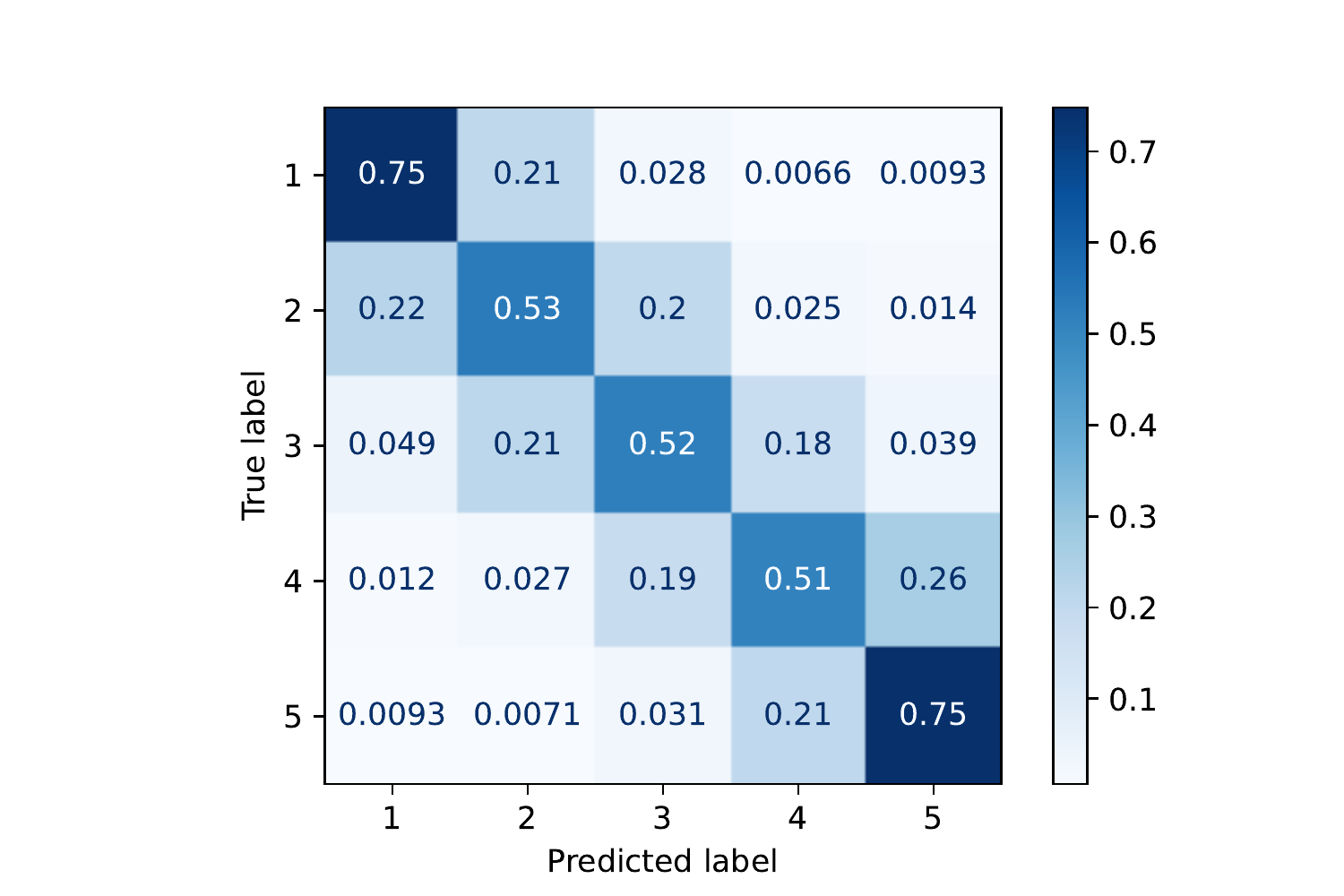}
	\caption{Confusion Matrix for Logistic Regression on the Testing Set}
	\label{fig 4-1}
\end{figure}

The second applied machine learning model is Logistic Regression \cite{ESL} (\texttt{LogisticRegression}) with $ L^2 $ penalty and LBFGS solver. The accuracy and weighted $ F_1 $ score on the testing set are shown in the Table \ref{tab 4-1}, and the confusion matrix is shown in the Figure \ref{fig 4-1}. Logistic Regression model outperforms the Naive Bayes with 64\% accuracy, and this result matches reported one \cite{elkouri2015predicting}.

%-----------------------------------------------------------
\subsubsection{Random Forest}

\begin{figure}[!ht]
	\centering
	\includegraphics[width=\linewidth]{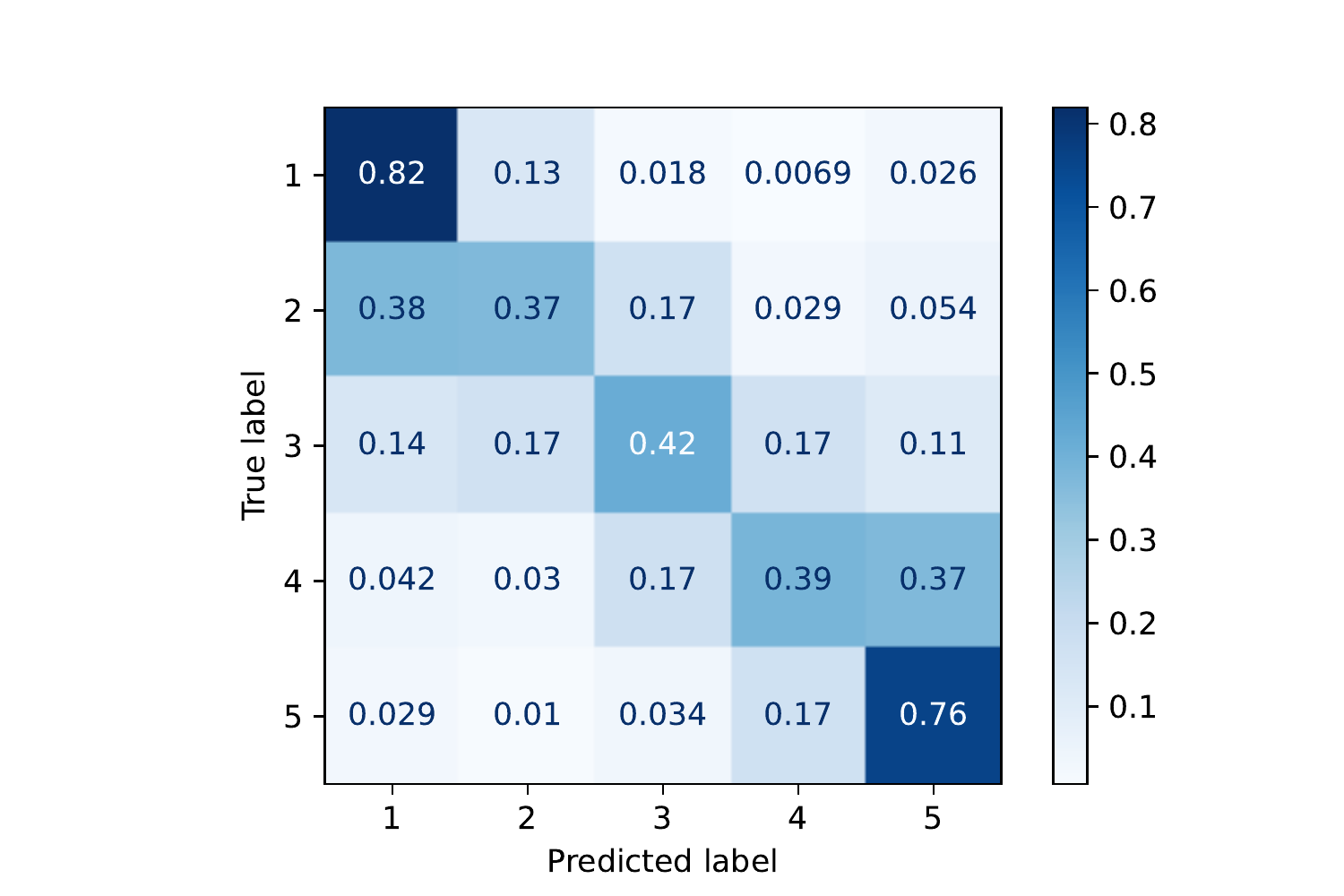}
	\caption{Confusion Matrix for Random Forest on the Testing Set}
	\label{fig 4-2}
\end{figure}

The third applied machine learning model is Random Forest \cite{ESL} (\texttt{RandomForestClassifier}) with 500 estimators, minimum samples leaf 10, and Gini impurity. The accuracy and weighted $ F_1 $ score on the testing set are shown in the Table \ref{tab 4-1} with the confusion matrix in the Figure \ref{fig 4-2}. The prediction is skew towards two polarities. This is because both 1 and 2 stars reviews have negative sentiment words, and 4 and 5 stars reviews share positive ones. Decision tree based model would be clumsy to distinct polar cases. The top positive and negative words grouped by restaurant type can be found in \cite{yu2017identifying}.

%-----------------------------------------------------------
\subsubsection{Linear Support Vector Machine}

\begin{figure}[!ht]
	\centering
	\includegraphics[width=\linewidth]{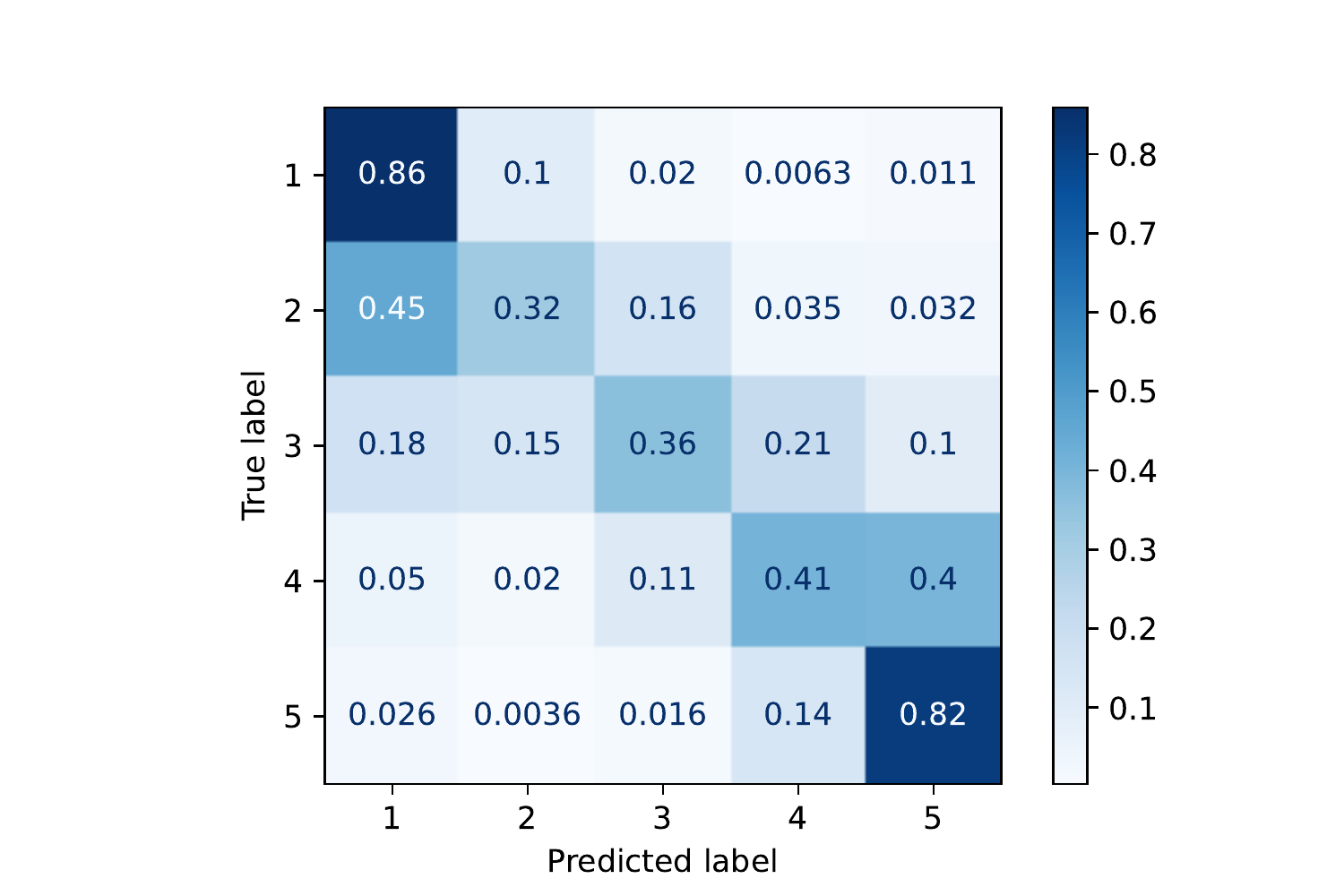}
	\caption{Confusion Matrix for Linear Support Vector Machine on the Testing Set}
	\label{fig 4-3}
\end{figure}

The fourth applied machine learning model is Linear SVM \cite{ESL} (\texttt{SGDClassifier}, linear classifier with stochastic gradient descent training) with hinge loss and $ L^2 $ penalty,  where \texttt{StandardScaler} is used for feature preprocessing. The accuracy and weighted $ F_1 $ score on the testing set are shown in the Table \ref{tab 4-1}, and the confusion matrix is shown in the Figure \ref{fig 4-3}. Similar polarity skewness as Random Forest can also be found in this linear SVM model, although the accuracy is higher than the Naive Bayes.

%-----------------------------------------------------------
\subsection{Transformer-Based Models}

In recent years, transformer-based models \cite{Attention} become the important Natural Language Processing (NLP) landscapes, and they outperforms many other state-of-the-art methods on enormous tasks. In the second part of experiment, four transformer-based models are used on our Yelp review rating prediction task, including BERT \cite{BERT}, DistilBERT \cite{DistilBERT}, RoBERTa \cite{RoBERTa}, and XLNet \cite{XLNet}. 

One NLP Python package \href{https://github.com/ThilinaRajapakse/simpletransformers}{\texttt{simpletransformers}} \cite{simpletransformers} based on \texttt{transformers} \cite{Transformers} and \texttt{pytorch} \cite{PyTorch} is used in this section. 

%-----------------------------------------------------------
\subsubsection{BERT}

The first applied transformer-based model is BERT. BERT (Bidirectional Encoder Representations) \cite{BERT} is a language representation model, where deep bidirectional representation from unlabeled text is pre-trained on both left and right context for masked language model and next sentence prediction tasks. Such pre-trained BERT model can used for wide tasks after being fined-tuned with just one additional output layer.

Two BERT architectures are available: one smaller model called base BERT with 110M parameters (12 transformer blocks, 768 hidden size, 12 self-attention heads), and another larger model called large BERT with 340M parameters (24 transformer blocks, 1024 hidden size, 16 self-attention heads). Both of them can be uncased and cased, depending on the pretraining English text. We use \texttt{bert-base-uncased}, \texttt{bert-base-cased}, and \texttt{bert-large-cased}, and their results on the validation set are shown in the Table \ref{tab 4-4}. It can be found that the cased, large models outperform the uncased, based ones. This conforms to the usual expectations, since upper-case words could have different meanings than lower-case ones, and lager model has more parameters to fit the data. However, large BERT model only gives 0.31\% increase in accuracy with double running time. To keep a reasonable running time on Google Colab, we will use base architectures for other transformer-based models.

\begin{table}[!ht]
	\centering
	\caption{Evaluation for Pretrained Cased Base and Large BERT Architectures on the Validation Set with 128 Maximum Sequence Level}
	%	(Max Seq. Level: maximum sequence level, $ F_1 $: weighted $  F_1 $ score)
	\vspace{4pt}
	\begin{tabular}{ccc}
		\toprule
		Architecture & Accuracy & $ F_1 $ Score \\
		\midrule
		\texttt{bert-base-uncased} & 0.6898 & 0.6947 \\
		\texttt{bert-base-cased} & 0.6961 & 0.6999 \\
		\texttt{bert-large-cased} & 0.6992 & 0.7034 \\
		\bottomrule
	\end{tabular}
	\label{tab 4-4}
\end{table}

For the transformer-based model, one of hyperparameters to be determined is the maximum sequence level. Base BERT models pretrained on the cased English text with maximum sequence level 32, 64, and 128 are chosen here for simplicity and representativeness. The results on the validation set are show in the Table \ref{tab 4-3}, where one can find the larger maximum sequence level is, the higher accuracy, since longer sequences can contain more useful information for classification. Although shorter sequence length will give worse results, it is still meaningful when memory resources are limited and faster computation is needed. Since the maximum sequence level with 256 will be out of memory on Google Colab, we will use 128 as the maximum sequence level for other transformer-based models.

\begin{table}[!ht]
	\centering
	\caption{Evaluation for Different Maximum Sequence Levels with Cased Base BERT on the Validation Set}
	%	(Max Seq. Level: maximum sequence level, $ F_1 $: weighted $  F_1 $ score)
	\vspace{4pt}
	\begin{tabular}{ccc}
		\toprule
		Max Sequence Level & Accuracy & $ F_1 $ Score \\
		\midrule
		32 & 0.6008 & 0.6090 \\
		64 & 0.6581 & 0.6638 \\
		128 & 0.6961 & 0.6999 \\
		\bottomrule
	\end{tabular}
	\label{tab 4-3}
\end{table}

\begin{figure}[!ht]
	\centering
	\includegraphics[width=\linewidth]{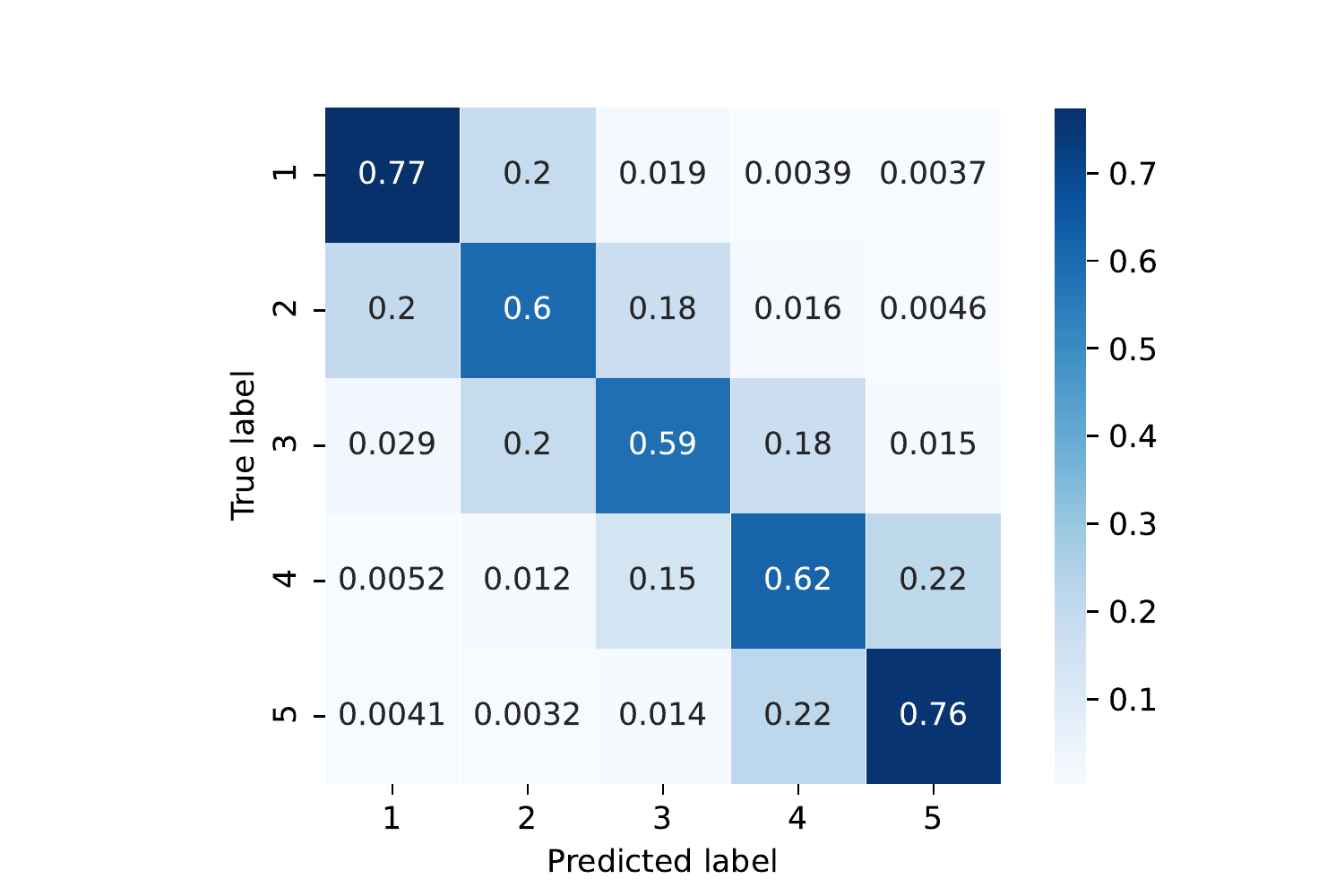}
	\caption{Confusion Matrix for Uncased Base BERT on the Testing Set}
	\label{fig 4-4}
\end{figure}

\begin{figure}[!ht]
	\centering
	\includegraphics[width=\linewidth]{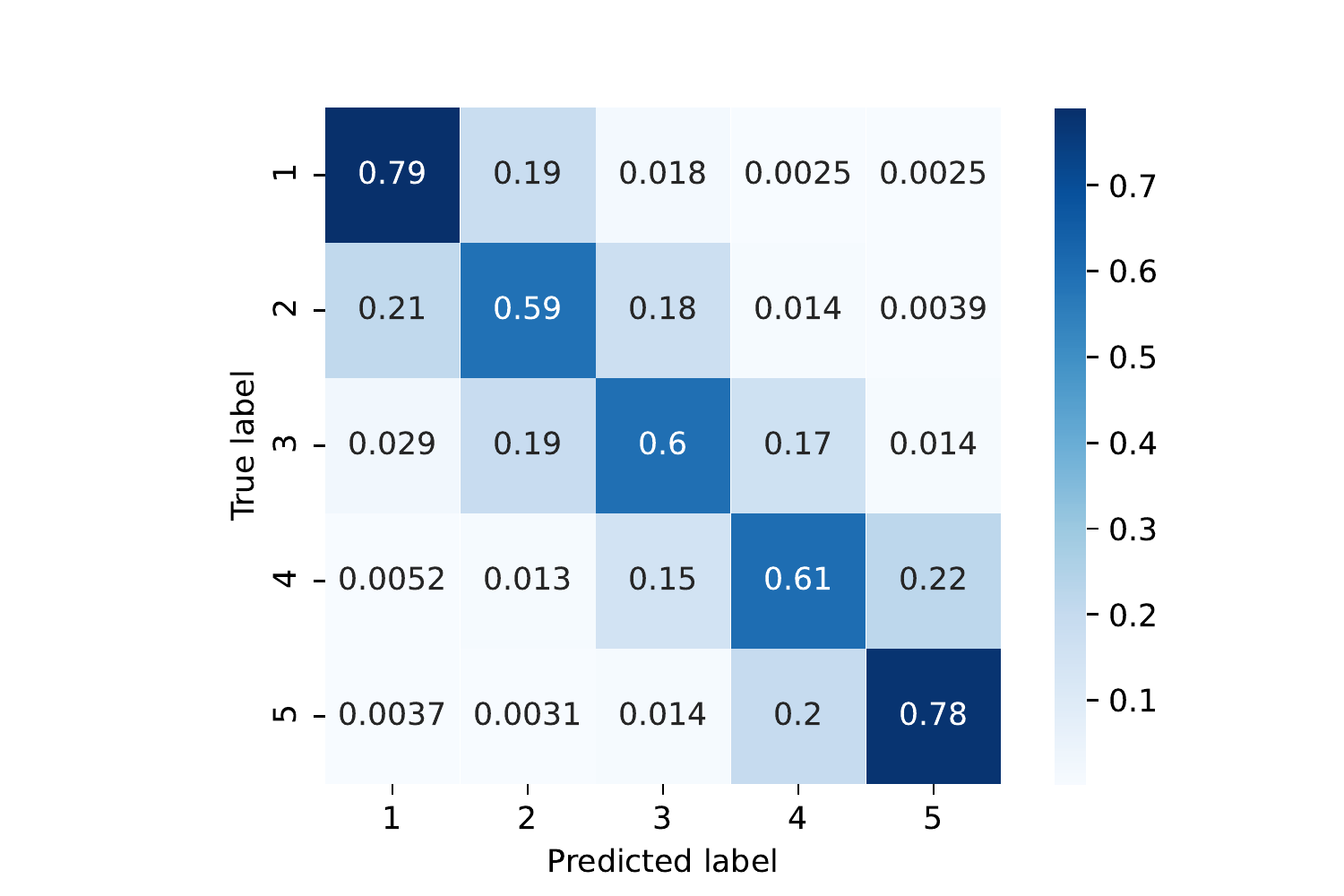}
	\caption{Confusion Matrix for Cased Base BERT on the Testing Set}
	\label{fig 4-8}
\end{figure}

In summary, we apply BERT models with following settings:
\begin{itemize}
	\item Architecture: base model pretrained on uncased (lower-case) or cased English text,
	\item Maximum sequence length: 128,
	\item Training batch size: 16,
	\item Number of training epoch: 1.
\end{itemize}
The accuracy and weighted $ F_1 $ scores on the testing set are shown in the Table \ref{tab 4-1}. The corresponding confusion matrices are shown in the Figure \ref{fig 4-4} and \ref{fig 4-8}. By comparing the results from base BERT models with the Logistic Regression, we can find the following outcomes:
\begin{itemize}
	\item The overall accuracy increases from 64\% to 69\%.
	\item The weighted $ F_1 $ score increases from 0.64 to 0.70.
	\item The BERT models have good abilities to distinct categories between 2-4 stars, which gives higher scores on diagonals of confusion matrices.
\end{itemize}

\begin{figure}[!ht]
	\centering
	\includegraphics[width=\linewidth]{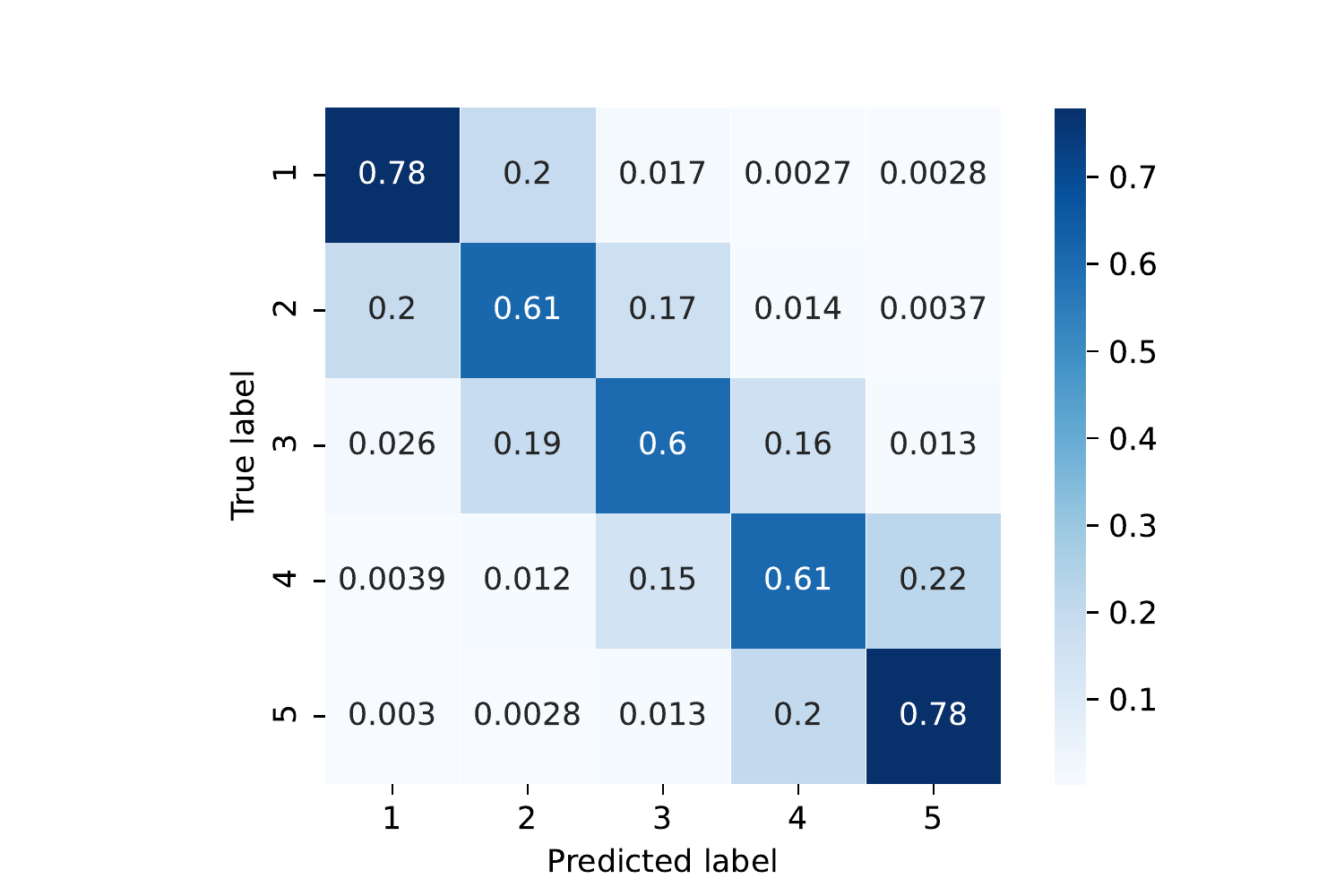}
	\caption{Confusion Matrix for Cased Large BERT on the Testing Set}
	\label{fig 4-10}
\end{figure}

%-----------------------------------------------------------
\subsubsection{DistilBERT}

The second applied transformer-based model is DistilBERT. DistilBERT (Distilled BERT) \cite{DistilBERT} is a distilled version of BERT, where 97\% language understanding capabilities of BERT are retained with 60\% faster performance and 40\% smaller model size.

\begin{figure}[!ht]
	\centering
	\includegraphics[width=\linewidth]{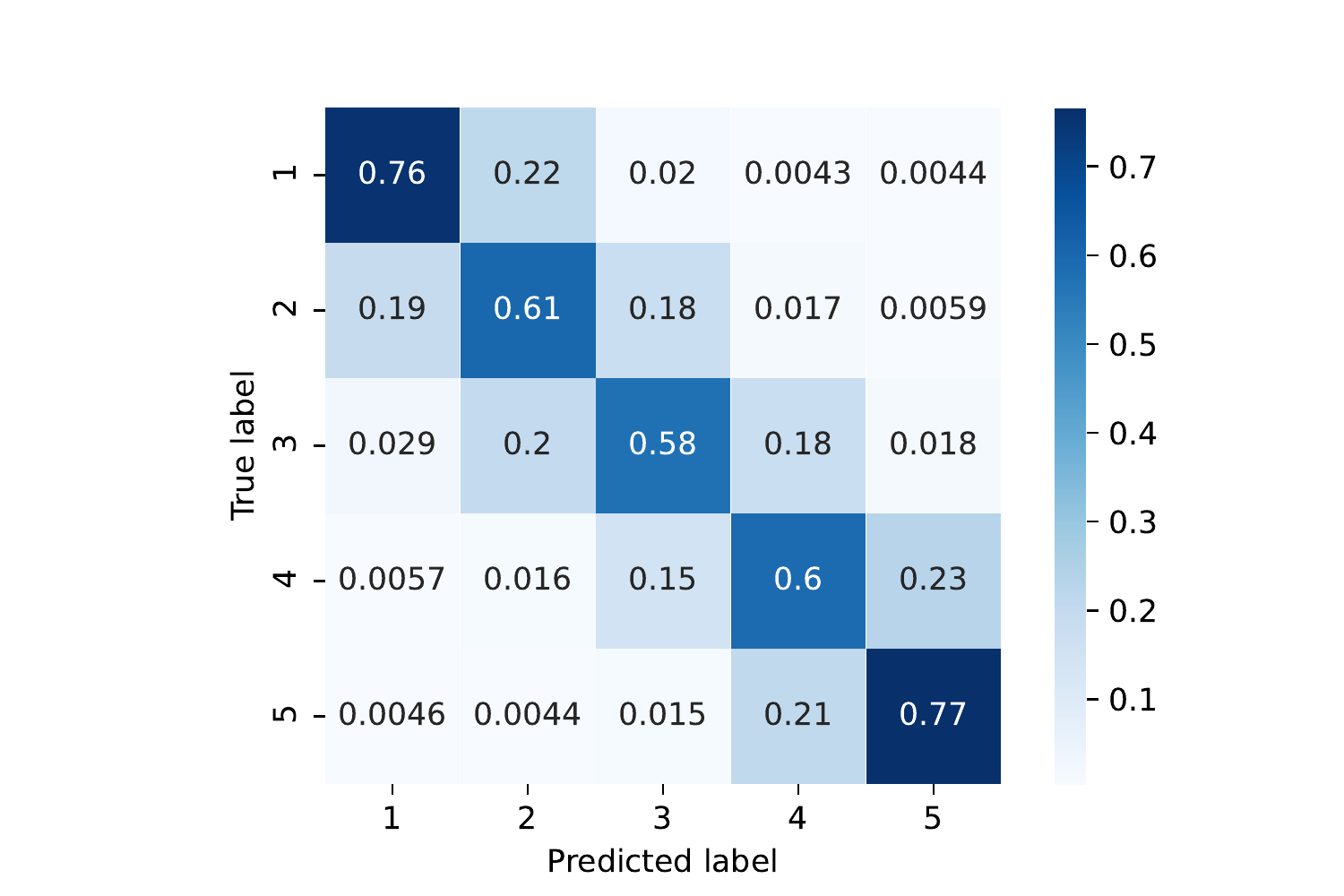}
	\caption{Confusion Matrix for Uncased Base DistilBERT on the Testing Set}
	\label{fig 4-5}
\end{figure}

\begin{figure}[!ht]
	\centering
	\includegraphics[width=\linewidth]{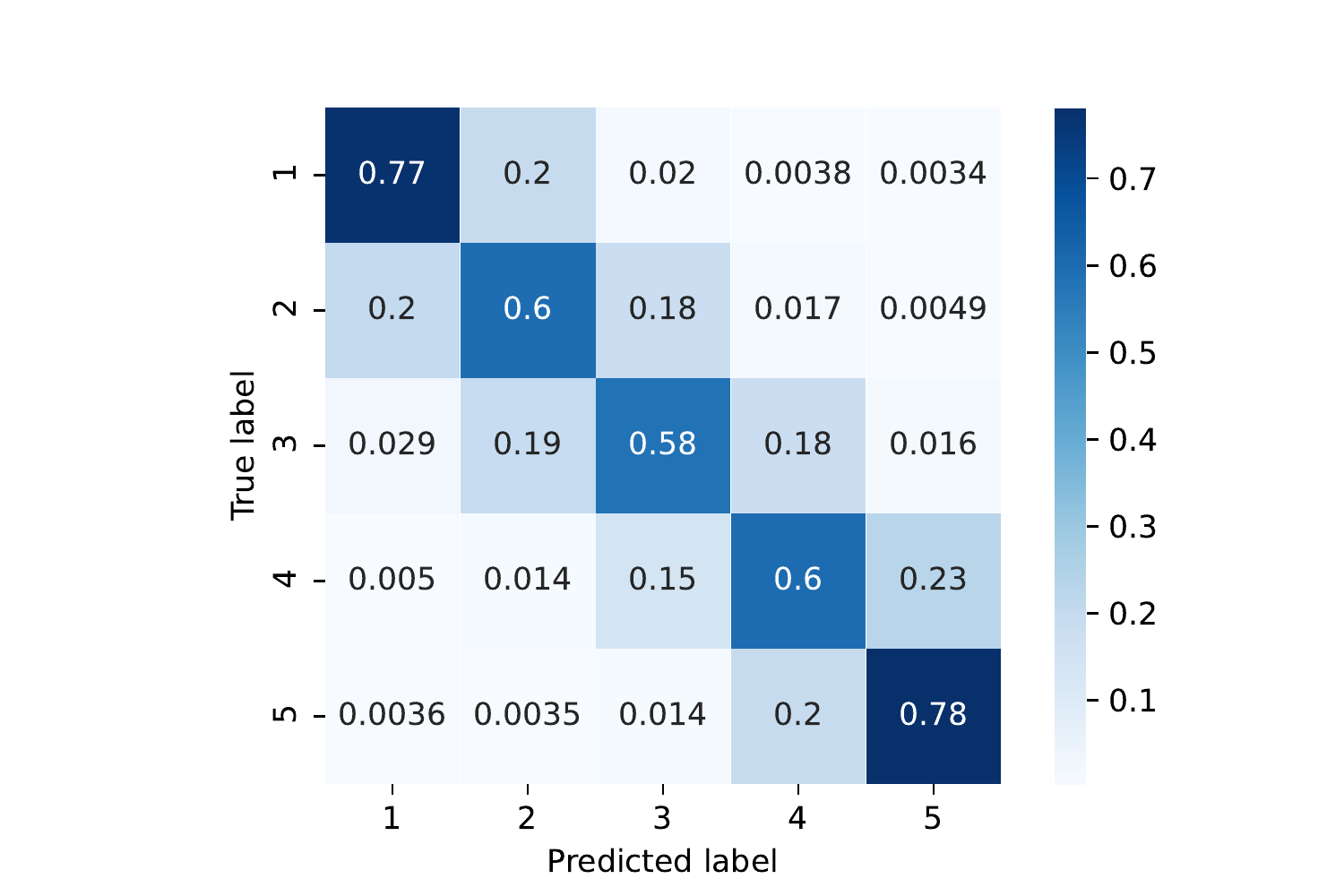}
	\caption{Confusion Matrix for Cased Base DistilBERT on the Testing Set}
	\label{fig 4-9}
\end{figure}

In this multiclass classification project, we use pretrained \texttt{distilbert-base-uncased} and \texttt{distilbert-base-cased} architectures. The accuracy and weighted $ F_1 $ scores on the testing set are shown in the Table \ref{tab 4-1} with corresponding confusion matrices in the Figure \ref{fig 4-5} and \ref{fig 4-9}. Compared with BERT, DistilBERT model has lower accuracy in around 0.5\% with almost double speed up. When the computation resources are limited, DistilBERT will be an appealing choice.

%-----------------------------------------------------------

\subsubsection{RoBERTa}

The third applied transformer-based model is RoBERTa. RoBERTa (Robustly optimized BERT approach) \cite{RoBERTa} improves pretraining procedure and achieves state-of-the-art results on several NLP tasks.

\begin{figure}[!ht]
	\centering
	\includegraphics[width=\linewidth]{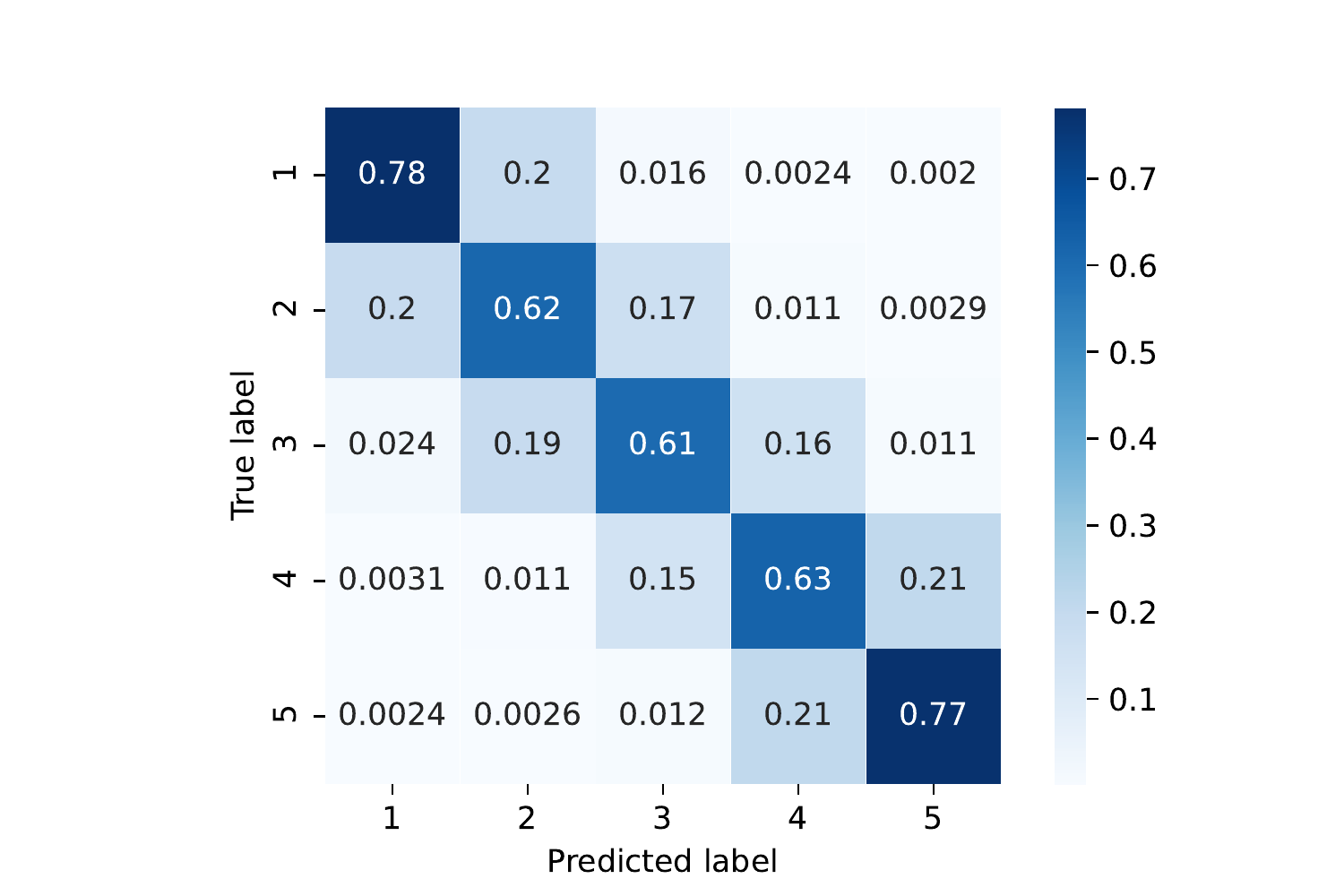}
	\caption{Confusion Matrix for Base RoBERTa on the Testing Set}
	\label{fig 4-6}
\end{figure}

For this project, we use pretrained \texttt{roberta-base} architecture. The accuracy and weighted $ F_1 $ score on the testing set are shown in the Table \ref{tab 4-1}. The confusion matrix is also shown in the Figure \ref{fig 4-6}. Higher scores are achieved as expected compared with the BERT model. A base RoBERTa even outperforms a large BERT model with comparable running time. This suggests the RoBERTa model will be a good choice for higher prediction metrics with sufficient computation computational resources.

%-----------------------------------------------------------

\subsubsection{XLNet}

The last applied transformer-based model is XLNet. XLNet \cite{XLNet} is a generalized auto-regressive pre-training method improved from BERT. In XLNet, all possible permutations for tokens are predicted, while only masked tokens in BERT. Also, ideas from Transformer-XL are integrated into XLNet. These allows XLNet outperforming BERT on 20 NLP tasks.

\begin{figure}[!ht]
	\centering
	\includegraphics[width=\linewidth]{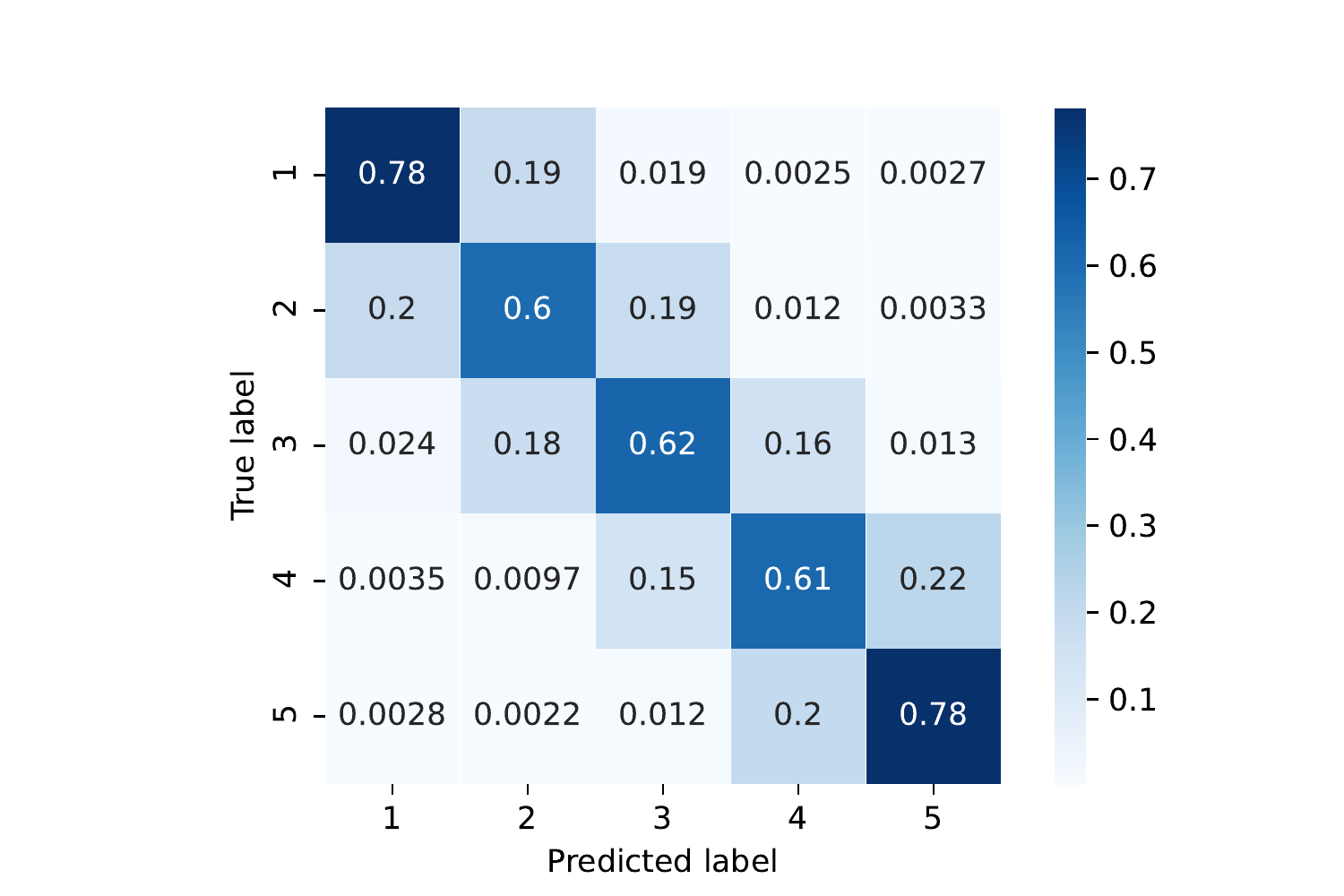}
	\caption{Confusion Matrix for Cased Base XLNet on the Testing Set}
	\label{fig 4-7}
\end{figure}

In this classification project, we use pretrained \texttt{xlnet-base-cased} architecture. The accuracy and weighted $ F_1 $ score on the testing set are shown in the Table \ref{tab 4-1} with the confusion matrix in the Figure \ref{fig 4-7}. XLNet model gives the best performance over all models we tried, also the longest running time. The 70.44\% accuracy is achieved on the testing set.

%-------------------------------------------------------------------------------
\section{Conclusions}

In this paper, we predicted ratings from Yelp review texts. Yelp Open Dataset was used. The imbalanced data distribution was presented, and a balanced training dataset was built. Four machine learning models including Naive Bayes, Logistic Regression, Random Forest, and Linear Support Vector Machine were used based on numerical features from tf-idf vectorizer. Four transformer-based models including BERT, DistilBERT, RoBERTa, and XLNet were also trained and tested on the textual data. Comparisons between models and hyperparameters were done, and 64\% accuracy score for the machine learning model and 70\% accuracy score for the transformer-based one were achieved on the testing set. 

Transformer-based models were summarized and experimented. Cased, large BERT models were found giving better performances than the uncased, base ones. DistilBERT has a faster computation speed with a bit lower metrics, while RoBERTa and XLNet give higher evaluation metrics with more computational resources required.

We hope our work could give some inspirations and insights for further work in Yelp review rating prediction based on machine learning and deep learning models.

%-----------------------------------------------------------
{\small
\bibliographystyle{ieee_fullname}
\bibliography{egbib}
}
\end{document}